\documentclass[english]{cccconf}
\usepackage[comma,numbers,square,sort&compress]{natbib}
\usepackage{amsmath}
\usepackage{multirow}
\begin{document}

\title{Sentiment Analysis of Online Travel Reviews \protect\\Based on Capsule Network and Sentiment Lexicon }
\author{Jia Wang, 
        Junping Du*,
        Yingxia Shao, and
        Ang Li}

\affiliation{School of Computer Science (National Pilot School of Software Engineering), Beijing University of Posts and Telecommunications, Beijing Key Laboratory of Intelligent Telecommunication Software and Multimedia, P.~R.~China.
        }
\maketitle

\begin{abstract}
With the development of online travel services, it has great application prospects to timely mine users' evaluation emotions for travel services and use them as indicators to guide the improvement of online travel service quality. In this paper, we study the text sentiment classification of online travel reviews based on social media online comments and propose the SCCL model based on capsule network and sentiment lexicon. SCCL model aims at the lack of consideration of local features and emotional semantic features of the text in the language model that can efficiently extract text context features like BERT and GRU. Then make the following improvements to their shortcomings. On the one hand, based on BERT-BiGRU, the capsule network is introduced to extract local features while retaining good context features. On the other hand, the sentiment lexicon is introduced to extract the emotional sequence of the text to provide richer emotional semantic features for the model. To enhance the universality of the sentiment lexicon, the improved SO-PMI algorithm based on TF-IDF is used to expand the lexicon, so that the lexicon can also perform well in the field of online travel reviews.
\end{abstract}

\keywords{Capsule Network, Sentiment Lexicon, Travel Reviews, BERT}

\footnotetext{*Corresponding author: Junping Du (junpingdu@126.com).}

\section{Introduction}
It is not difficult to see that the Internet has been deeply integrated into people's daily life. More and more people choose to release their emotions on social networks\cite{review1,Kou16Social,Li22Scientific,Yang15Ontology}. Therefore, how to use of Internet public opinion tendency to guide actual production and life has become a very important application prospect of text sentiment analysis. By introducing sentiment analysis technology, exploring online travel reviews can help online travel enterprises analyze customer needs promptly, to make timely feedback on their shortcomings, which has considerable application value.

With the brilliant performance of the BERT pre-training model in the NLP field, more and more attention has been paid to the downstream task of text sentiment classification using a hybrid network integrating Convolutional Neural Network (CNN) and Recurrent Neural Network (RNN)
\cite{review2}. Despite the continuous progress of the RNN model structure, the application of the traditional CNN structure in the NLP field seems to need to be improved. What's more, these network structures can extract context features from text sequences pretty well, but they still retain the disadvantage of sentiment text classification tasks based on the deep learning method, that is, they ignore the semantic features of the text.

In this paper, we proposed a text sentiment classification model based on Capsule Network and sentiment lexicon (SCCL) to analyze online travel reviews. We mainly improve the shortcomings of BERT-BiGRU from two aspects. On the one hand, the convolution unit is introduced to enhance the learning of local features; on the other hand, the sentiment lexicon is used to provide more sufficient semantic features for the model. However, this method entails some substantial challenges. First, when traditional CNN is applied to NLP tasks, due to the limitations of convolution kernel size and application of pooling layer, a large number of spatial features are lost. Directly adding it to the model will affect the excellent context information brought by BERT and BiGRU. Second, the generalization of the sentiment lexicon is very poor. Besides, it is time-consuming and laborious to build a lexicon fitting to the field manually, and such a lexicon has a strong subjectivity, which will greatly affect the final effect of the classification model. In order to solve these two problems, we take Weibo comments as an example and use capsule units instead of traditional convolution neurons. Meanwhile, manually select sentiment seed words to expand the domain lexicon based on the SO-PMI algorithm to obtain a sentiment lexicon suitable for Weibo comments. Then we verified the effectiveness of our method on the labeled Weibo comment public data set and compared it with the BERT classification model. Finally, we crawled the travel related comments on the social platform through the crawler, screened and manually marked them, so that we completed the sentiment classification of online travel comments.

The contribution of this paper lies in 2 aspects:
\begin{itemize}
  \item The capsule convolution cell is introduced to the BERT-BiGRU model instead of the traditional convolution neuron, so that the model can extract the local features of the text without destroying the good context features

  \item The domain lexicon is extended based on the SO-PMI algorithm with the Weibo comment data set, and the extended domain lexicon is used to provide semantic features for the deep learning classification model.
\end{itemize}

\section{Related Work}
Sentiment classification is the basis of sentiment analysis. Traditional text classification only pays attention to the objective content of the text, while sentiment classification studies more about subjective factors of the text author, that is, the emotional tendency of the expression. Generally, the research on text sentiment classification is classified from three technical levels: the method based on sentiment lexicon, the method based on machine learning, and the method based on deep learning.

\subsection{The Method Based on Sentiment Lexicon}
The sentiment classification method based on the sentiment lexicon mainly 
matches the words used in the text with the lexicon, then analyzes the overall emotional tendency of the text by processing the set of sentiment words that hit the lexicon. Song et al. use the sentiment lexicon to quantify the emotional intensity of the text at the word level
\cite{senti1}.
According to the number of words hitting the emotional dictionary and the weight of pre-related adverbs, they get the overall emotional score of the text. Yang et al. mined the emotional tendency of video comments through the SO-PMI algorithm and summarized and sorted out the sentiment lexicon in the field of video comments based on the standard sentiment lexicon\cite{senti2}.
Zhang et al. use the sentiment lexicon to analyze the emotion of video barrage and expand the lexicon through the word2vec algorithm\cite{senti3}.
Although it can efficiently recognize sentiment patterns with certain accuracy only by matching the lexicon quickly, this method does not consider the relationship between words, that is, there is no context in the analysis, which leads to a poor generalization of this method. The emotional value of a single word could not change dynamically according to the article or sentence. A word has different meanings in different times, different languages, and different cultures\cite{review3}. In particular, as complex linguistics, Chinese has countless polysemy phenomena. At the same time, the effect of this method also depends heavily on the quality of the lexicon. If you want to maintain a high level of the model, you need to maintain the dictionary quite often. Therefore, there are fewer and fewer methods to use lexicon alone.

\subsection{The Method Based on Machine Learning}
The sentiment classification method based on machine learning mainly focuses on K-Nearest Neighbor \cite{Sun09Study}, Naive Bayesian, and Support vector machines\cite{ml1, ml2}. Compared with the method based on sentiment lexicon, the machine learning model does not rely on manual construction, which reduces subjectivity. The classification model can be updated in time through the database. But this method usually cannot make full use of the contextual information of contextual text in emotional analysis, which will affect the accuracy. Therefore, the method based on deep learning has become the mainstream method.

\subsection{The Method Based on Deep Learning}
Deep learning is a subset of machine learning and an application of multi-layer neural networks in learning \cite{Fang20Identity,Li17consensus,Xu13Image,Lin09Average,Meng16seeking,Li14LPV}. For the problem of sentiment classification, there are two crucial network structures: CNN and RNN.

CNN is usually used in image processing, but the research using CNN as text feature extraction has gradually increased recently. Shao combined Bert and TextCNN models and verified the effectiveness of the method through the takeaway review dataset\cite{cnn1}. Xu et al. constructed a hierarchical classification model pos-ACNN-CNN by introducing location coding and attention mechanism into the ordinary CNN model and achieved good results on the IMDB film review dataset\cite{cnn2}.

LSTM has been proposed and widely used in text sentiment classification to solve the problem of gradient vanishing when RNN deals with long-distance dependence. As the optimized version of LSTM, Gated Recurrent Unit (GRU) model combines the forgetting gate and input gate in LSTM into an update gate, which reduces one-third of the parameters and has a faster iteration speed. Nowadays, bi-directional models such as BiLSTM and BiGRU have emerged, because these models are more in line with the characteristics of text information related to both front and back sequences.Yue et al. realized sentiment analysis through the BiLSTM network, before the word vector entered the neural network, they first increased the weight of the word vector of keywords through the pre-attention mechanism\cite{rnn1}. Compared with the traditional model, the accuracy can be improved by 1.7\%. Lu et al. built an improved BiGRU that can adapt to the recursive network structure\cite{rnn2}. In addition, Textblob technology is used to correct spelling errors during preprocessing, so that the model can well avoid errors caused by spelling errors.  These studies bring more semantic features to the methods based on deep learning. 

At first, people thought that each CNN and RNN corresponds to different learning tasks, so a single network model is usually used to extract text features \cite{Shi19collaborative,Li13Region,Li17Kalman,Zhao17sliding,Hu18Anomaly,Li17estimation,Li13Gaussian}. Later, people gradually realized that the hybrid CNN and RNN model can extract text information more comprehensively. Yan et al. constructed a dual-channel feature extraction model of CNN and BiGRU compared the performance of different pre-training word embedding models, and proved the effectiveness of introducing the attention mechanism into the hybrid model\cite{mix1}. Deng et al. used a network integrating CNN and RNN\cite{mix2}. On the one hand, the model improved the extraction of local information from the text, on the other hand, it well preserved the context relevance of the text. Finally, the attention mechanism was used to enhance the role of words with strong emotional tendencies in classification.
Li et al. introduced the genetic algorithm on the basis of the CNN-LSTM hybrid network to help the model better search for the optimal solution globally\cite{mix3}.

However, due to the size of the convolution kernel and the application of the pooling layer, the traditional CNN model is insensitive to the spatial information of features, which is very unfriendly to text processing. Therefore, some scholars also use the Temporal Convolution Network (TCN) which is conducive to the processing of text structure\cite{tcn1, tcn2}. Due to its unique causal convolution structure, TCN is a unidirectional model, which makes it not in line with the trend of using a bidirectional structure in the NLP field. Therefore, some researchers have used spatial information-sensitive Capsule Convolution Networks to deal with emotional text classification.Based on using CNN and BiGRU as feature extraction networks, Cheng et al. introduced capsule structure as a classifier to classify text emotion\cite{cap1}. 
Xu has achieved good performance in small sample data sets and cross-domain migration by integrating attention mechanism and capsule structure, using a multi-head self-attention-based feature extractor and capsule structure as a classifier\cite{cap2}.
Zhang used the capsule network to overcome the disadvantage of information loss in convolution neural network pooling operation and uses the capsule structure to replace the traditional convolution structure in the BiGRU-CNN model\cite{cap3, cap4}.

Recently, some researchers have proposed sentiment text classification combining the sentiment lexicon method and deep learning method.
Luo et al. combined affective dictionary and deep learning methods and introduced sentiment lexicon features and residual network structure into the Bi-LSTM network structure. On the one hand, the compilation volume of the sentiment lexicon was reduced through LSTM, and on the other hand, the model performance was improved compared with standard LSTM and BERT\cite{tol2}.
Duan et al. proposed a hierarchical classifier based on BERT and adaptive sentiment lexicon, which uses sentiment lexicon processing to improve the classification accuracy for the classification probability after the BERT model classification\cite{tol3}.
Yang et al. mainly applied sentiment lexicon to word embedding, preliminarily screened text words through sentiment lexicon, then used char CNN to embed words, and finally used the ELMo model to classify text\cite{tol5}.
 However, all these methods either lack the consideration of local features or lack the semantic feature information, so they cannot extract the text features comprehensively.

\begin{figure}[!htb]
  \centering
  \includegraphics[width=\hsize]{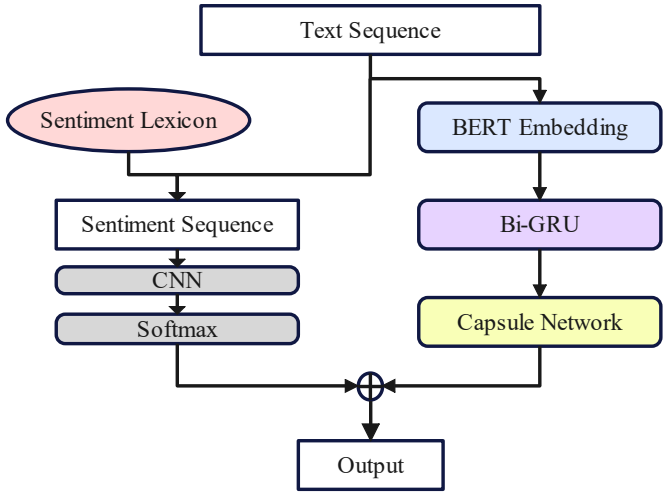}
  \caption{Overall structure of SCCL}
  \label{model}
\end{figure}

\section{Model}
This section will describe the overall situation of the SCCL model and introduce the key algorithms in the model in detail.

\subsection{Overview}
Fig.~\ref{model} shows a simple block diagram of the overall structure of our SCCL model. It can be seen that the SCCL model is generally divided into two routes to extract text features. On the one hand, the left half of Fig.~\ref{model} shows the sentiment semantic feature extraction route. First, the text sequence is segmented, and the stopped words are removed, then the sentiment lexicon is used to match the text emotional feature words, the hit emotional feature words are combined into emotional sequences and embedded through the word2vec algorithm, then the emotional semantic features are extracted through the feature extraction network of CNN and Softmax. On the other hand, in the text context feature extraction part shown in the right half of Fig.~\ref{model}, instead of word segmentation, Bert is directly used for character-level embedding, and then the combined network of BiGRU and Capsule network is used to extract text context features. Finally, the model will synthesize the results of the two parts of feature extraction and integrate the output.

\begin{figure}[!htb]
  \centering
  \includegraphics[width=\hsize]{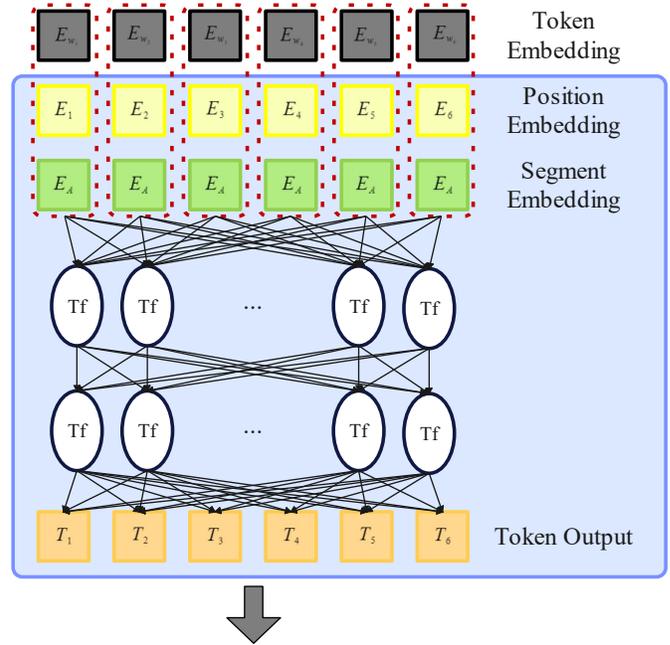}
  \caption{Simple structure of BERT}
  \label{bert}
\end{figure}

\subsection{BERT for Embedding}
The full name of Bert is bidirectional encoder representation from transformers, which is a pre-training language representation model based on the Transformers model. Fig.~\ref{bert} shows the simple structure of BERT and $ Tf $ modules represent Transformers block. It no longer adopts the traditional one-way language model like the GPT model or performs pre-training after shallow splicing of two one-way language models like the ELMo model, but adopts a new Masked Language Model (MLM), so as to generate deep bi-directional language representation. At the same time, BERT will embed each text sequence in three ways when inputting text sequences, namely token embeddings, segment embeddings, and position embeddings, which is also an indispensable feature for Bert to better extract context information.
Since BERT was proposed, it has achieved overwhelming advantages in many NLP tasks, and even aroused the pre-training frenzy in this field. Using BERT pre-training model has almost become the only choice for text processing.

\subsection{BiGRU}
GRU is a simplified and improved neural network model for LSTM. The LSTM module is composed of three gating units: input gate, forgetting gate, and output gate. In GRU neural network, the three gating units in LSTM are replaced by the update gate $ z_{t} $ and reset gate $ r_{t} $. In this way, the parameters and tensors of the model are reduced, making GRU more concise and efficient than LSTM. The structure of the GRU unit is shown in Fig.~\ref{model}. 
\begin{figure}[!htb]
  \centering
  \includegraphics[width=\hsize]{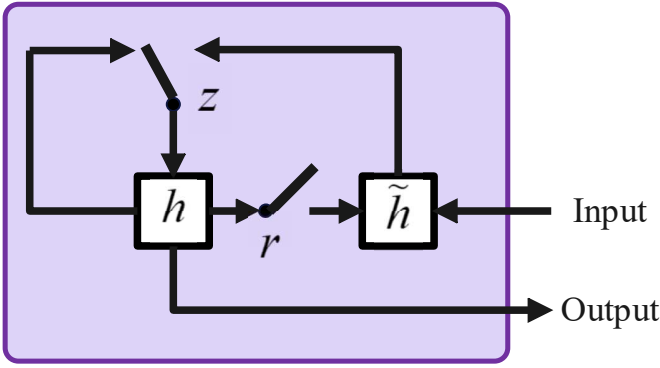}
  \caption{GRU unit}
  \label{gru}
\end{figure}

In the actual calculation process, GRU first obtains the gating information of $ z_{t} $ and $ r_{t} $ through the current input $ x_{t} $ and the hidden state $ {h}_{t-1} $ passed down from the previous node. The calculation formula is shown in formula~(\ref{gru1}) and formula~(\ref{gru2}), where $ \sigma $ is the Sigmoid activation function, and $ W_{z} $ and $ W_{r} $ are the weight parameters of the two gates respectively.
\begin{equation}
  \label{gru1}
    z_{t} = \sigma (W_{z}({h}_{t-1}, x_{t}))
\end{equation}

\begin{equation}
  \label{gru2}
    r_{t} = \sigma (W_{r}({h}_{t-1}, x_{t}))
\end{equation}

After obtaining the gating information, the current input $ x_{t} $ is spliced with the reset gate, and then the output of the currently hidden node $ \Tilde{h}_{t} $ is activated by the tanh activation function. The calculation formula is shown in formula~(\ref{gru3}), where $ W $ is the weight parameter of the hidden layer.
\begin{equation}
  \label{gru3}
    \Tilde{h}_{t} = tanh (W(r_{t}{h}_{t-1}, x_{t}))
\end{equation}

Finally, the state of hidden layer $ {h}_{t} $ is updated according to the state of update gate $ {z}_{t} $.
\begin{equation}
  \label{gru4}
    h_{t} = (1 - z_{t})h_{t-1} + z_{t}\Tilde{h}_{t}
\end{equation}

In the classical GRU, the transmission of the state is one-way from front to back. However, in some problems, the output at the current time is related not only to the previous state but also to the subsequent state. For example, judging the emotional polarity of a polysemous word requires not only the previous judgment but also the content of the following text. The emergence of BiGRU solves this problem. The SCCL model used the BiGRU network to learn global semantic information from the input matrix $ X $. in the training process, the network uses two GRU models to model emotion along the forward and backward of the text sequence and outputs the hidden layer $ {H}_{t} $. The specific calculation process is shown in formula~(\ref{gru5}), formula~(\ref{gru6}) and formula~(\ref{gru7}).
\begin{equation}
  \label{gru5}
    h^{\rightarrow}_{t} = GRU(X, h^{\rightarrow}_{t-1}), t \in [1, L]
\end{equation}

\begin{equation}
  \label{gru6}
    h^{\leftarrow}_{t} = GRU(X, h^{\leftarrow}_{t-1}), t \in [L, 1]
\end{equation}

\begin{equation}
  \label{gru7}
    H_{t} = [h^{\rightarrow}_{t}, h^{\leftarrow}_{t}]
\end{equation}

Where $ h^{\rightarrow}_{t} \in R^{L \times d} $ is the emotional feature representation of word vector matrix $ X $ fused with the previous information, $ h^{\leftarrow}_{t} \in R^{L \times d} $ refers to the fusion of emotional features later, and $ d $ refers to the output vector dimension of GRU unit. $ H_{t} \in R^{L \times 2d} $ combines the two in series to fuse the contextual emotional information as the feature of the input text at the BiGRU layer.

\subsection{Capsule Network}
As an excellent feature detector, CNN can use a one-dimensional convolution kernel to extract local patterns from vector sequences, and then use the pooling operation to reduce the number of parameters, so as to complete the extraction of text features. However, these characteristics of CNN have also become an obstacle for the model to make an important breakthrough in the text field. On the one hand, the size of the convolution kernel determines the feature size that CNN can detect. For text processing, too large a convolution kernel will make it difficult for the model to learn the relationship between words, while too small a convolution kernel will make the model unable to deal with complex sentence structures such as inversion and preposition. On the other hand, the pooling operation will cause the CNN structure to lose a lot of spatial information. In image processing, a pooling operation can help CNN detect important pixel information in the picture while reducing parameters. Even if some spatial information is lost, it can summarize the picture features according to important pixels. However, the text sequence is more sensitive to space, especially when the complex Chinese is the processing target, the pooling operation is easy to ignore the details hidden in the text, resulting in errors. In contrast, the capsule network cannot be limited by the structure and size of the detection unit, and can automatically update the receptive domain, so that it can flexibly grasp the internal spatial relationship between the whole and local text, to learn the complex internal relationship of the text. 

\begin{figure}[!htb]
  \centering
  \includegraphics[width=\hsize]{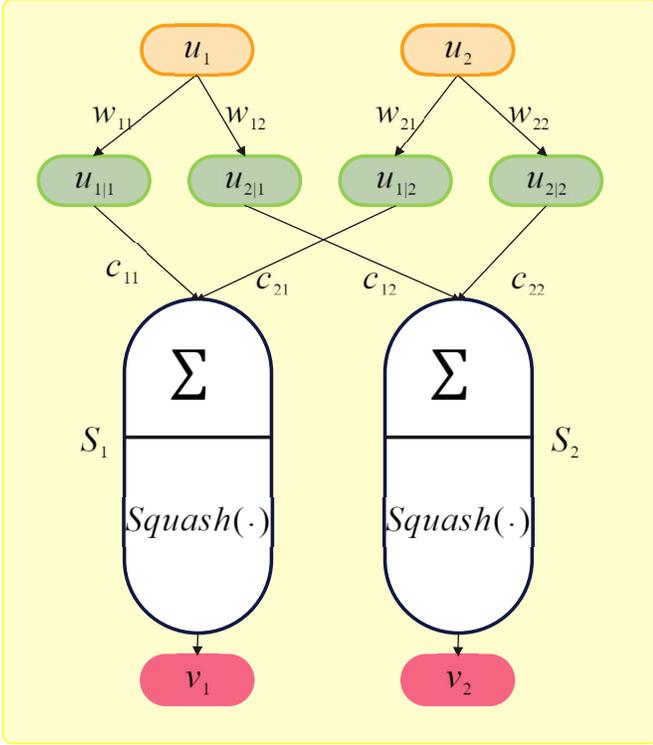}
  \caption{Capsule layer structure}
  \label{cap}
\end{figure}

The core idea of the capsule network is to replace the traditional convolution neuron with the capsule layer. However, the capsule layer is composed of neurons, except that ordinary neurons accept a series of scalars and output scalars, while the full connection layer of the capsule layer accepts a series of vectors and outputs vectors. Therefore, the internal structure of the capsule layer is also different from that of ordinary neurons. In the capsule layer structure shown in in Fig.~\ref{cap}, the primary capsule layer will divide the feature vector into multiple primary capsules through simple channel cutting. After the primary capsules enter the routing capsule layer, they first encode the position information through the affine transformation as formula~(\ref{capeq1}). In th
is formula, $ u_{i} $ represents the $ i^{th} $ primary capsule and $ \bar{u}_{j|i} $ represents one of its high layer capsule, $ W_{ij} $ shows the affine metrics acting on it.
\begin{equation}
  \label{capeq1}
    \bar{u}_{j|i} = W_{ij}u_{i}
\end{equation}

Then, the capsule network will dynamically route the high-level capsules according to formula~(\ref{capeq2}). In short, the dynamic routing is a weighted sum based on the weight $ c_{ij} $. The calculation method of the weight $ c_{ij} $ is shown in formula~(\ref{capeq3}). $ c_{ij} $ determines how the vector after the affine matrix projection processing of this layer will enter the vector of the next layer. Unlike traditional neurons, the dynamic routing does not require bias.
\begin{equation}
  \label{capeq2}
    S_{j} = \sum_{i}C_{ij}\bar{u}_{j|i}
\end{equation}
\begin{equation}
  \label{capeq3}
    C_{ij} = \frac{exp(b_{ij})}{\sum_{k}exp(b_{ik})}
\end{equation}

It is not difficult to see that the key to dynamic routing is how to determine the weight coefficient $ c_{ij} $. For ordinary neurons using maximum pooling, only one value can enter the next layer, that is, when $ c_{ij} $ is a one-hot vector. The capsule layer uses a method called Routing-by-agreement, which gives weight to the vector in a similar clustering way. This process is completed by the Softmax operation on the prior probability $ b_{ij} $ from the low-level capsule i to the high-level capsule j. In each space after projection, the weight of the more clustered vectors will increase, and vice versa. In this way, the main features of all vectors can be extracted.

In addition to dynamic routing, another technical point of the capsule network is the use of the unique activation function Squash shown in formula~(\ref{capeq4}).
\begin{equation}
  \label{capeq4}
    v_{j} = \frac{{||s_j||}^2}{1+{||s_j||}^2}\frac{s_j}{||s_j||}
\end{equation}

It can be seen that Squash is a normalization operation, which takes each vector between 0 and 1 without affecting the direction. In formula~(\ref{capeq4}), the former term represents the probability that the features contained in the vector are perceived by the high-level network, and the latter term is the unit vector that maintains the direction.

\subsection{Expanding Domain Lexicon Based on SO-PMI}
The existing sentiment lexicon including this lexicon can only recognize a limited number of affective words. At the same time, Chinese words have polysemy and fuzziness. The meaning of the same word in different fields may be different, especially for text sentiment analysis in the social media environment. On the one hand, because there are many new words in the text in the social media environment, the traditional sentiment lexicon is difficult to detect new words; On the other hand, the text content of social media has strong domain characteristics, and the general sentiment lexicon may bring serious misjudgment of information. Therefore, the emotional words in different fields should not be the same. We must build a unique domain sentiment lexicon for the field of Weibo comment text.

In order to identify the special emotion words in the field of online comments, this paper uses the combination of TF-IDF and SO-PMI to expand the sentiment lexicon. Compared with the traditional SO-PMI, this method can consider more semantic information of emotional words and reduce manpower\cite{senti2}. TF-IDF algorithm is a statistical method to evaluate the importance of a word relative to a certain text in the whole corus. PMI is used to measure the relevance between two words. The calculation formula is shown in formula~(\ref{pmieq}), which $ P(w_{1}, w_{2}) $ shows the probability of word $ w_{1}$ and $ w_{2}$ in the same text and $ P(w_{1}) $ and $ P(w_{2}) $ represent the occurrence probability of $ w_{1}$ and $ w_{2}$ respectively. The SO-PMI algorithm judges whether a word is more likely to appear with positive words or negative words based on the PMI value of words. The calculation formula is shown in formula~(\ref{sopmieq}). 
\begin{equation}
  \label{pmieq}
    PMI(w_{1}, w_{2}) = log_2[\frac{P(w_{1}, w_{2})}{P(w_{1})P(w_{2})}]
\end{equation}

\begin{equation}
    \label{sopmieq}
    \begin{split}
        SO-PMI(word) &= \sum_{seed \in Pos}PMI(word, seed) \\
        &- \sum_{seed \in Neg}PMI(word, seed)
    \end{split}
\end{equation}
We first rank the words in the corpus according to their TF-IDF values, and then manually select the 100 most important high-frequency emotional words. The 100 high-frequency emotion words include 50 positive emotion words and 50 negative emotion words represented by $ Pos $ and $ Neg $ in formula~(\ref{sopmieq}). Then the 100 emotion words are used as emotion seed words, and then the SO-PMI values of the words in the corpus are calculated to select 60 positives and 60 negatives with the most obvious emotional tendency, which are added to the sentiment to form a domain lexicon.

\section{Experiment}
This section mainly introduces the results of the experiment based on our SCCL model, including the used text datasets and sentiment lexicon, then briefly analyzes the experimental results.

\subsection{Dataset}
The text data set we used in this study for model training is from the public Weibo Review data set. This data set contains 40133 texts, which are divided into six categories. These data comes from 17133 items in the NLPCC Emotion Classification Challenge and 23000 labeled Weibo review data. During the training, 10\% of the training data will be randomly selected as the test set to test the performance of the model. Then we use the crawler technology to crawl a number of travel comments on Weibo and other social platforms, filter and manually mark 2000 comments as the target set, we ensure that the category distribution is equivalent to the training set and the total number is equivalent to the testing set in this process. The classification of data sets is shown in Tables~\ref{dataset}. 

\begin{table}[!htb]
  \centering
  \caption{Classification of data sets}
  \label{dataset}
  \begin{tabular}{c| c c c}
    \hhline
    Dataset & Class & Label & Number \\ \hline
    \multirow{6}{*}{training} 
    & Null & 0 & 13993 \\ 
    & Like & 1 & 6697 \\ 
    & Sad & 2 & 5348 \\ 
    & Disgust & 3 & 5978 \\ 
    & Anger & 4 & 3167 \\ 
    & Happiness & 5 & 4950 \\
    \hline
    \multirow{6}{*}{target}
    & Null & 0 & 700 \\ 
    & Like & 1 & 200 \\ 
    & Sad & 2 & 300 \\ 
    & Disgust & 3 & 300 \\ 
    & Anger & 4 & 200 \\ 
    & Happiness & 5 & 300 \\
    \hhline
  \end{tabular}
\end{table}

The collection of emotion words and their related emotion resources is called the sentiment lexicon. It is an important supporting resource for text sentiment analysis and mining\cite{senti5}. HowNet's Chinese sentiment lexicon is used in our experiment, which has 3730 Chinese positive evaluation words, 3116 Chinese negative evaluation words, 836 Chinese positive emotion words, and 1254 Chinese negative emotion words.

\subsection{Effectiveness of Capsule Network}
In order to prove that the Capsule network can perform well in text processing, we replace the capsule structure in the SCCL model with a multi-head CNN for comparison. At the same time, in order to verify whether other convolution structures sensitive to time and space can also surpass the performance of traditional CNN, we also try to use the Temporal Convolution Network TCN to replace the Capsule network. Causal Convolution, an important structure of TCN, is a convolution model proposed to deal with sequence problems. This convolution structure can combine the information of pre-sequence states when predicting a certain state, which gives the network time sensitivity. On the other hand, another important structure of TCN, Dilated Convolution can expand the receptive field of the convolution mechanism, which endows the model with spatial sensitivity. The experimental results are shown in Tables~\ref{effect1}.
\begin{table}[!htb]
  \centering
  \caption{Text sequence feature extraction network}
  \label{effect1}
  \begin{tabular}{l|c c}
    \hhline
    Model & acc & f1 \\ \hline
    CNN & 48.68 & 46.59 \\ 
    Multi-head CNN & 51.11 & 49.72 \\ 
    TCN & 48.92 & 47.47 \\ 
    \textbf{Capsule Network} & \textbf{52.45} & \textbf{51.0} \\
    \hhline
  \end{tabular}
\end{table}

It can be seen that the overall performance of the Capsule structure is better than that of the traditional CNN and TCN, which also stems from the fact that the Capsule structure can better retain the space-time information of the feature vector and ensure that the features will not be lost in the deep network. On the contrary, although the traditional CNN has expanded the receptive fields as much as possible through the multi-head mechanism, the convolution results of connecting different receptive fields in parallel are not flexible enough, and the effect is not as good as the Capsule network. In addition, although TCN overcomes the disadvantage of space-time insensitivity, the prediction of this convolution model in a certain state cannot be related to the information at any future time, which also violates the trend of NLP using bi-directional structure. In particular, when it is used together with BERT and BiGRU, the bi-directional features obtained by the low-level network are lost to a certain extent, resulting in a decline in performance.

\subsection{Effectiveness of Sentiment Lexicon}
In order to choose which feature extraction network to use to extract the features of sentiment sequences, we also conducted a group of comparative experiments. Because the sentiment sequence is only composed of some intermittent emotional words, the context feature is weaker than the text sequence, or we don't want the model to learn wrong context information through it. Therefore, we don't use the feature extraction structure similar to the text but only use the simple convolution extraction structure. At the same time, we replace the simple convolution in SCCL with multi-head convolution, capsule convolution, and temporal convolution for comparison. The results are shown in Tables~\ref{effect2}. It can be seen that the effects of several convolution structures are similar, but the simple CNN structure still has better performance.
\begin{table}[!htb]
  \centering
  \caption{Sentiment sequence feature extraction network}
  \label{effect2}
  \begin{tabular}{c|c c}
    \hhline
    Model & acc & f1 \\ \hline
    Multi-head CNN & 52.09 & 49.6 \\ 
    TCN & 52.01 & 49.9 \\ 
    Capsule Network & 52.27 & 51.13 \\ 
    \textbf{CNN} & \textbf{52.45} & \textbf{51.0} \\
    \hhline
  \end{tabular}
\end{table}

\subsection{Ablation Experiment}
Finally, in order to verify the effectiveness of parts of SCCL, we also conducted ablation experiments. The main method is to remove all parts of the SCCL model or replace them with other structures. The results are shown in Table 1. It can be seen that the SCCL model proposed in this paper is indeed better than the network of BERT direct classification or BERT words embedded in BiGRU to extract features. At the same time, the introduction of a Capsule network can bring greater improvement than the introduction of the sentiment lexicon, which is mainly limited by the quality of the sentiment lexicon. In particular, the sentiment lexicon we use only contains the emotional polarity of words, but the actual classification task has six categories. Positive emotion and negative emotion are divided into two categories and three categories respectively, which also limits the accuracy of classification. Nevertheless, the effect of using the lexicon with domain expansion is better than that of the common lexicon and the method without a lexicon. At the same time, the SCCL model can also achieve 48\% accuracy on our target set, which proves that our model has certain generalization.
\begin{table}[!htb]
  \centering
  \caption{Ablation experiment results}
  \label{effect3}
  \begin{tabular}{l|c c}
    \hhline
    Model & acc & f1\\ \hline
    BERT & 50.04 & 47.94 \\ 
    BERT-BiGRU & 51.05 & 49.14 \\
    BERT-CapsuleNet & 50.75 & 48.86 \\
    BERT-BiGRU-CapsuleNet & 51.63 & 49.98 \\
    BERT-BiGRU-Normal Lexicon & 51.00 & 48.86 \\
    BERT-BiGRU-Expanded Lexicon & 51.15 & 49.43 \\
    \textbf{SCCL} & \textbf{52.45} & \textbf{51.0} \\ 
    \textbf{SCCL-test} & \textbf{48.58} & \textbf{47.43} \\ 
    \hhline
  \end{tabular}
\end{table}

\section{Conclusion}
In this paper, we study the text sentiment classification model SCCL based on capsule network and sentiment lexicon from two aspects: how to enhance the local feature extraction ability of the mixed language model and how to provide more emotional semantic features for the sentiment classification model. Through experimental verification, it is proved that the capsule network as a local feature extractor is better than the traditional convolutional neural network. It is also found that TCN and the bi-directional language model are not effective when used together. In addition, we also confirmed that the introduction of a sentiment lexicon into the deep learning model can add better semantic features to the model to improve performance, and this dictionary must be expanded in combination with the application field, otherwise, it is difficult to get excellent results.

As future work, we plan to start with building a better domain lexicon. In this experiment, the sentiment lexicon we use can only distinguish the positive and negative polarity of emotional words, but we are facing a problem of multiple classifications. Both positive and negative texts are divided into different categories, which makes it difficult for the lexicon to play a role in distinguishing these texts.

\end{document}